\documentclass[sigconf, 9pt, nonacm]{acmart}

\AtBeginDocument{%
  }

\usepackage{float}
\usepackage{amsmath,amssymb,amsfonts}
\usepackage{graphicx,color}
\usepackage{textcomp}
\usepackage[utf8]{inputenc} 
\usepackage[T1]{fontenc}    
\usepackage{hyperref}       
\usepackage{url}            
\usepackage{booktabs}       
\usepackage{nicefrac}       
\usepackage{microtype}      
\usepackage{subcaption}
\usepackage{graphicx}
\usepackage{algorithm}
\usepackage{algorithmic}
\usepackage{afterpage}
\usepackage{wrapfig}
\usepackage{svg}
\begin{document}


\title{Non Line-of-Sight Optical Wireless Communication using Neuromorphic Cameras}

\author{Abbaas Alif Mohamed Nishar}
\affiliation{
  \institution{Georgia State University}
  \city{Atlanta}
  \state{Georgia}
  \country{USA}
  \postcode{30324}}
\email{amohamednishar1@student.gsu.edu}

\author{Alireza Marefat}
\affiliation{
  \institution{Georgia State University}
  \city{Atlanta}
  \state{Georgia}
  \country{USA}
  \postcode{30324}}
\email{amarefatvayghani1@student.gsu.edu}

\author{Ashwin Ashok}
\affiliation{
  \institution{Georgia State University}
  \city{Atlanta}
  \state{Georgia}
  \country{USA}
  \postcode{30324}}
\email{aashok@gsu.edu}
\begin{abstract}
Neuromorphic or event cameras, inspired by biological vision systems, capture changes in illumination with high temporal resolution and efficiency, producing streams of events rather than traditional images. In this paper, we explore the use of neuromorphic cameras for passive optical wireless communication (OWC), leveraging their asynchronous detection of illumination changes to decode data transmitted through reflections of light from objects. We propose a novel system that utilizes neuromorphic cameras for passive visible light communication (VLC), extending the concept to Non Line-of-Sight (NLoS) scenarios through passive reflections from everyday objects. Our experiments demonstrate the feasibility and advantages of using neuromorphic cameras for VLC, characterizing the performance of various modulation schemes, including traditional On-Off Keying (OOK) and advanced N-pulse modulation. We introduce an adaptive N-pulse modulation scheme that dynamically adjusts encoding based on the packet's bit composition, achieving higher data rates and robustness in different scenarios. Our results show that lighter-colored, glossy objects are better for NLoS communication, while larger objects and those with matte finishes experience higher error rates due to multipath reflections. 

\end{abstract}

\ccsdesc[500]{General and reference~Cross-computing tools and techniques}
\ccsdesc[300]{Information systems~Signal processing}
\ccsdesc[500]{Hardware~Communication hardware, interfaces and storage}
\ccsdesc[300]{Computer systems organization~Sensor networks}
\ccsdesc[500]{Applied computing~Optical communications}
\ccsdesc[300]{Computer systems organization~Neural networks}
\ccsdesc[300]{Hardware~Emerging optical and photonic technologies}
\ccsdesc[500]{Computing methodologies~Computer vision}

\keywords{Neuromorphic cameras, Optical wireless communication (OWC), Non-line-of-sight (NLoS), Pulse shift keying (PSK), Adaptive modulation, N-pulse modulation, Computer vision}

\pagenumbering{gobble}
\maketitle

\section{Introduction}
Neuromorphic processing and neuromorphic cameras have emerged as groundbreaking technologies in artificial intelligence and computer vision\cite{2024spikingneuralnetworksfastmoving}\cite{neuro}. These cameras operate on principles inspired by biological systems, capturing changes in illumination with high temporal resolution and efficiency. Unlike conventional cameras, which capture frames at fixed rates, neuromorphic cameras asynchronously detect and record only significant changes in the scene, producing streams of events rather than images. This unique ability positions them as ideal candidates for innovative applications in various fields, including optical wireless communication (OWC) including where direct Line-of-Sight (LoS) may not be available.

OWC leverages the broad optical spectrum, including visible, ultraviolet, and infrared light, for data transmission. Conventional (non event) cameras, however, have limitations such as low frame rates, high energy consumption, and increased noise levels. The receivers that are predominantly used in OWC are cameras and photodiodes. Cameras are associated with substantial energy consumption, whereas photodiodes require amplification and precise optical lensing, which can introduce hardware complications. In addition, cameras often struggle with noise, particularly in signal isolation from the environment. Photodiodes, when operating with reverse bias, experience an amplified dark current, which can degrade the signal-to-noise ratio, especially in low-light environments~\cite{meetoptics_photodiodes}. Neuromorphic cameras offer a promising alternative as they can rapidly detect small changes in light intensity. Their ability to capture high frequency light modulations makes them ideal for OWC. 
\begin{table}[t!]
\centering
\scalebox{0.8}{%
\begin{tabular}{lccc}
\toprule
\textbf{Receiver Type} & \textbf{Sampling Rate} & \textbf{Dynamic Range} & \textbf{Resolution} \\
\midrule
Photodiode        & $\sim$1 GHz            & $\sim$100 dB         & Single-point       \\
High-Speed Camera & $\sim$1 kHz            & $\sim$70 dB          & 1920 $\times$ 1080  \\
Neuromorphic Camera & 20 kHz                 & 120 dB               & 1280 $\times$ 720  \\
\bottomrule
\end{tabular}
}
\caption{Comparison of Receiver Technologies for VLC}
\vspace{-0.25in}
\label{tab:receiver_specs}
\end{table}

Passive communication in VLC primarily leverages retro-reflective surfaces and digital micro-mirror devices (DMDs). The high dynamic range of event cameras provides us the ability to leverage indirect paths, such as reflections from surfaces, to enable information transfer even when a direct path between transmitter and receiver is obstructed. This feature is especially advantageous in cluttered and complex indoor environments where direct optical links may be blocked.

In this work, {\bf we propose a novel system that leverages neuromorphic cameras for passive Visible Light Communication (VLC) when the object and the camera are static}. We extend this concept to Non-Line-of-Sight (NLoS) scenarios by leveraging passive reflections from ordinary objects in an indoor setting, restructuring event data into \emph{event frames} via computer vision techniques, and evaluating the NLoS performance under different conditions. The contributions presented in this paper are as follows:

\begin{enumerate}
\item Demonstrating use of neuromorphic camera in static settings, enabled by events triggered by reflection of time-varying (VLC encoded) optical signals on the objects. 
\item Developing a system that uses surface reflections to enable NLoS optical wireless communication.
\item Characterizing the performance of NLoS VLC for various types of object sizes and surface finishes.
\end{enumerate}


\section{Related Works}
\textbf{Optical Camera Communication (OCC):} Aranda et al.~\cite{aranda2020npulse} introduced the use of event-based cameras for OCC, demonstrating how n-pulse modulation can be used to encode data efficiently, reducing the complexity of the demodulation and optimizing performance. While this work focuses on Line-of-Sight (LoS) usage with lower fidelity our work focuses on NLoS VLC in the aspects of Joint Communication and Sensing. Furthermore, our proposed adaptive N-pulse modulation scheme dynamically adjusts the encoding based on the packet's bit composition, optimizing both data rate and the robustness of communication. We also demonstrate the strength of this strategy that reporesents a departure from conventional OOK or fixed pulse schemes, as it adapts to channel conditions and bit patterns, ensuring reliable performance even when operating under NLoS conditions with indirect reflections.

\vspace{1mm}\noindent\textbf{RF and Millimeter Wave Systems:} Recent studies, such as those of Fang et al.~\cite{fang2022joint}, have explored the integration of communication and sensing in RF and millimeter wave systems. These studies highlight the potential of using MIMO systems to achieve high precision in sensing while maintaining robust communication links.
Techniques utilizing WiFi signals for simultaneous communication and environmental sensing have also been investigated, employing variations in the strength and phase of the WiFi signal to detect movement and presence within an environment \cite{hrm}.

\vspace{1mm}\noindent\textbf{Event-Based VLC Systems:} Previous works such as those by Deng et al.~\cite{deng2023occreview} provide a comprehensive review of optical camera communication systems, including the use of neuromorphic cameras. Xu et al.~\cite{vlc_backscat} have discussed the implications of utilizing neuromorphic cameras in backscatter communication for vehicular networks. Yu et al.~\cite{remark} uses VLC backscatter communication with retroreflectors for single pixel imaging (SPI) which captures minimal information required for positioning and identifying markers. The SPI combined with retroreflectors further helps in capturing only necessary light signals further enhancing privacy, however the communication performance is subpar. 
Recent work by Wang et al.~\cite{wang2024highspeedpassivevisiblelight} have shown improvement in the data rate performance upto 1.6Mbps using mutiplexed multiple input multiple output (MIMO) channels using a digital micromirror device (DMD) for reflection. However, the complexity of the hardware setup using a DMD and that its optics requires precise placement and positioning of the camera relative to the object and under short ranges are practical limitations of this plausible yet restricted design. Our approach distinctly addresses the limitations of event cameras in static environments where we can use the events created by passive reflection to both sense and communicate in the environment for indoor scenarios. Event cameras inherently require changes in illumination to register events, hence without motion, no events are generated. Our work overcomes this by deliberately inducing light intensity variations using VLC, thereby enabling event generation even under zero-motion conditions. This capability not only extends the applicability of neuromorpohic cameras for NLoS VLC but also facilitates a dual modality - joint sensing and communication using the same hardware.

\vspace{1mm}\noindent\textbf{Computer Vision and Object Detection:} Event-based cameras have been widely used in applications such as obstacle detection in drones, as surveyed by Gehrig et al.~\cite{gehrig2022eventbased}. These studies highlight the ability of neuromorphic cameras to perform real-time object detection and tracking with low latency and high energy efficiency.

\vspace{1mm}\noindent\textbf{Scene Understanding and Navigation:} In robotics, neuromorphic cameras have facilitated advanced scene understanding and navigation. Neuromorphic cameras often pair well with spiking neural networks (SNN) which are particularly effective in processing the sparse and temporally rich data generated by event-based cameras, enabling sophisticated tasks~\cite{ziegler2024spiking} such as gesture recognition and predictive maintenance.

\section{Understanding Neuromorphic Cameras}
Neuromorphic cameras function by mimicking the human visual system. In the human eye, light passes through the cornea and pupil and is focused by the lens onto the retina, where photoreceptors convert light into electrical signals. These signals travel along the optic nerve to the brain, forming a perception of the environment. Similarly, neuromorphic sensors feature pixels that operate independently and asynchronously, reacting only to changes in the scene. This is in stark contrast to traditional frame-based sensors that integrate and accumulate photon energy over fixed intervals, often leading to redundant data. The fundamental difference in signal detection lies in the underlying physics. Traditional cameras integrate incident light over an exposure period, generating complete images with significant redundant information. However, Neuromorphic cameras detect relative changes in light intensity capturing only the dynamic aspects of a scene. This event-driven approach not only reduces data redundancy but also provides high temporal resolution (up to approximately 10,000 FPS) and low latency, both of which are crucial for high-fidelity VLC applications. Although high-speed cameras can capture thousands of frames per second, they produce voluminous data and may miss transient changes because of their fixed acquisition scheme. Photodiodes, while sensitive and fast, require additional amplification and precise optical alignment and are prone to increased dark current and noise under low-light conditions. The dual capability of neuromorphic cameras, to deliver high-fidelity communication and concurrently sense the environment, makes them particularly well suited for nonlinear-line-of-sight VLC systems. Table~\ref{tab:receiver_specs} contrasts the key specifications of photodiode, a high-speed camera, and a neuromorphic camera, highlighting the advanages of neuromorphic systesm in terms of dynamic range and a balanced sampling rate and resolution.

\begin{figure}[ht!]
    \centering
    \includegraphics[width=\columnwidth]{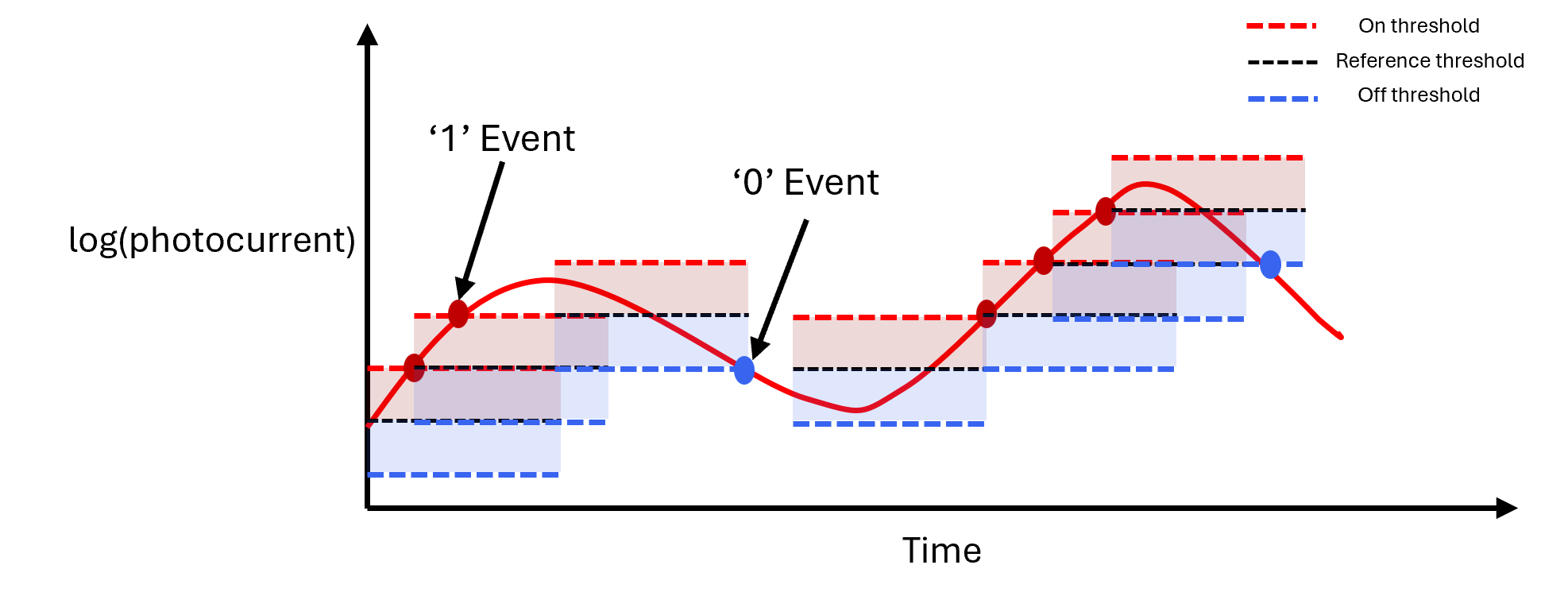}
    \caption{This diagram shows event generation in a neuromorphic camera based on light intensity changes. As the intensity crosses the positive threshold (Bias Diff On, red dashed line), an "on" event (red circle) is triggered. When it crosses the negative threshold (Bias Diff Off, blue dashed line), an "off" event (blue circle) is generated. Each pixel operates asynchronously, producing timestamped events only during changes, resulting in high temporal resolution, minimal data redundancy, and low latency.}  
    \label{fig:events_formation}
\end{figure}

\begin{figure}[ht!]
    \centering
    \includegraphics[width=\columnwidth]{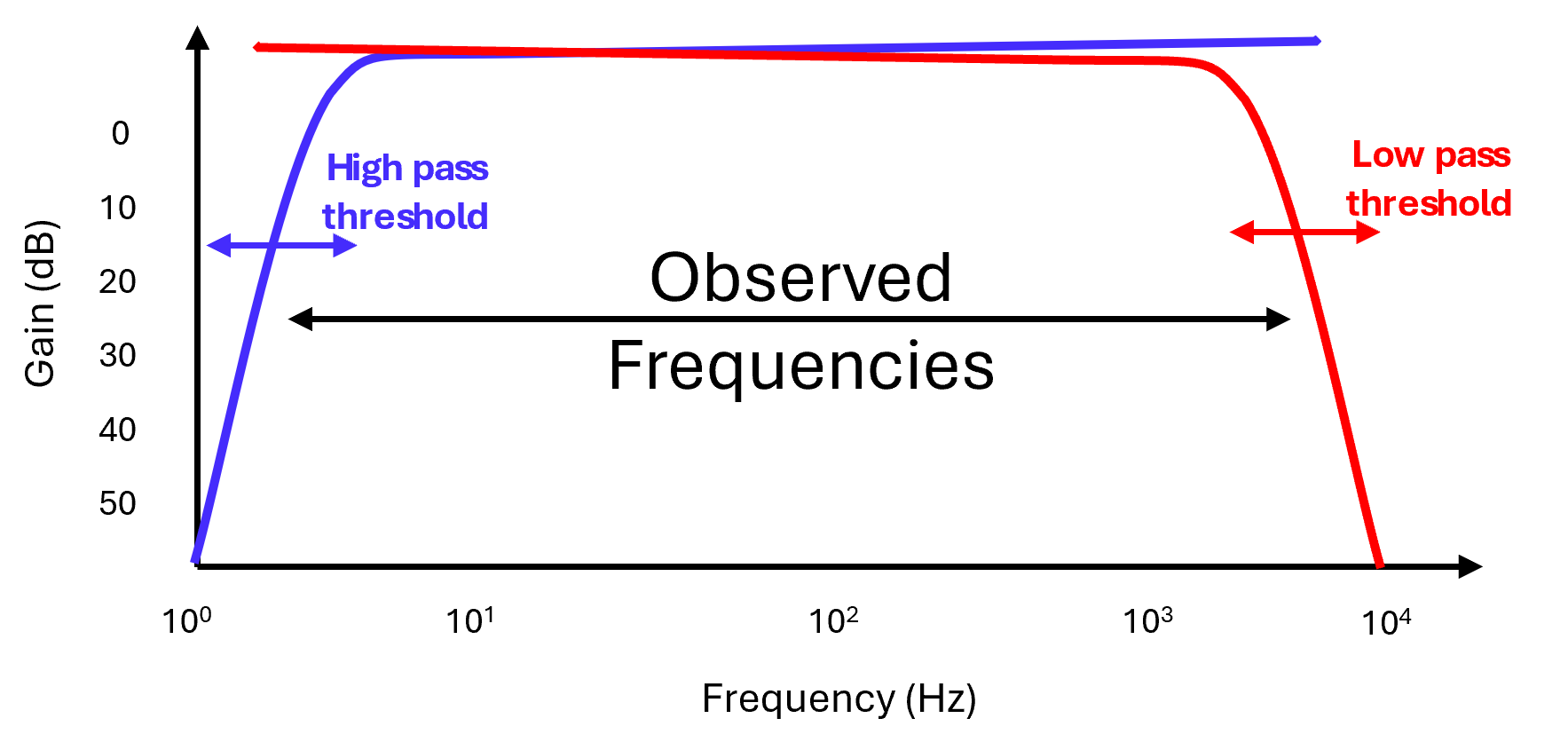}
    \caption{Spectral response of analog filters present in event cameras. The frequency filtered out by the high-pass and low-pass filters can be controlled by adjusting the bias values in the camera. The observed event rate is the band present in the overlap area between the high pass and low pass filters.}
    \label{fig:filter_neuro}
\end{figure}

\subsection{Event Generation}

The event-based operation implies that the sensor generates sparse data called events as reactions only when a significant change in light intensity is detected. Each pixel in an event-based sensor detects changes in temporal contrast autonomously and independently, as shown in Figure~\ref{fig:events_formation}. When a change in light intensity is detected, an event is generated and timestamped, providing continuous data that adapt dynamically to the scene's activity.
Specifically, when the intensity exceeds the positive threshold (Bias Diff On), an "on" event is produced; when it falls below the negative threshold (Bias Diff Off), an "off" event occurs. The sensor then updates its reference level to the new intensity, ensuring that only subsequent significant changes trigger further events. Each event is output with a timestamp, pixel coordinates, and polarity (1 or 0), with some systems (e.g., the IMX636 sensor with the EVK4 kit) achieving resolutions as fine as 26 microseconds~\cite{prophesee_evk4}.

The \emph{bias\_f0} and \emph{bias\_hpf} play a critical role in controlling the frequency range of events detected by the neuromorphic camera. \emph{bias\_f0} governs the low-pass filter, which affects the detection of high-frequency signals. Adjusting \emph{bias\_f0}, you can filter out fast-moving objects or flickering light, reducing high-frequency noise, but it can also increase pixel latency and event rate. In contrast, increasing \emph{bias\_f0} can decrease the latency of the pixel, allowing the sensor to capture high-frequency motion while potentially introducing more noise. 

\emph{bias\_hpf} controls the high-pass filter, which filters out low-frequency signals, such as slow-motion events or low-frequency noise. Increasing \emph{bias\_hpf} helps to remove these slow signals, effectively reducing the event rate and noise from low-frequency changes. Adjusting these biases allows for fine-tuning the sensor sensitivity and event rate based on the specific requirements of the application, as seen in Figure~\ref{fig:filter_neuro}, ensuring that only the most relevant events are captured while minimizing noise and data overload.

Some of the key Advantages of Neuromorphic Cameras are: 

(a) \textbf{High Temporal Resolution}: Neuromorphic cameras can capture rapid changes in the scene with minimal latency, making them ideal for joint sensing and communication in NLoS VLC

(b) \textbf{Low Data Redundancy}: By recording only changes in the scene, neuromorphic cameras generate significantly less data compared to traditional cameras, reducing the need for extensive data storage and processing.

(c) \textbf{Energy Efficiency}: The asynchronous nature of event-based sensors allows for lower power consumption, as the sensors only activate when a change is detected.

(d) \textbf{Dynamic Range}: Neuromorphic cameras can operate effectively in a wide range of lighting conditions, from low light to very bright environments, due to their high dynamic range.

\subsection{Event Sensors: EVK4 and IMX636}
Neuromorphic cameras, such as those that employ the IMX 636 sensor, are revolutionizing the field of visual data acquisition by emulating the functioning of the human eye. This section delves into the technical details and advantages of using the EVK4 system with the IMX 636 sensor, focusing on how these sensors work, their tuning processes, and their implications for optical wireless communication and joint sensing applications.

\vspace{1mm}\noindent{\bf Event-Based Sensor Functionality.}
Event-based sensors operate on a principle fundamentally different from traditional frame-based cameras. Instead of capturing complete images at regular intervals, they detect changes in the scene asynchronously, generating events only when a change in light intensity is detected. 

\vspace{1mm}\noindent{\bf Pixel Architecture and Event Generation.}
Each pixel in an event-based sensor works as an independent contrast change detector. When the light intensity at a pixel crosses a set threshold, an event is generated. This event includes a timestamp and the pixel's address in the sensor array, which can indicate either an increase (on-event) or decrease (off-event) in light intensity.

\vspace{1mm}\noindent{\bf Key Bias Settings and Their Impact.}
Adjusting the biases of an event-based sensor is crucial to optimize its performance. Key settings include: 

(i) Contrast Sensitivity Biases: Adjusting the thresholds for detecting changes in light intensity can make the sensor more or less sensitive, balancing between detecting subtle changes and minimizing noise. Adjusting the bias settings can significantly alter the sensor's sensitivity and noise levels, optimizing for different lighting conditions and application requirements;

(ii) Low-Pass and High-Pass Filters: These filters manage the frequency range of detected changes, which is crucial for filtering out noise or focusing on specific types of motion. Proper tuning of low-pass and high-pass filters can help in managing noise and improving the detection of relevant events;

(iii) Refractory Period Bias: This setting controls the dead time after an event during which no new events can be generated by the same pixel, affecting the overall event rate and temporal resolution. By modifying the refractory period, one can control the temporal resolution and event rate, which is critical for high-speed applications.

\vspace{1mm}\noindent{\bf Key Performance Indicators (KPIs).}
Important metrics for evaluating the performance of neuromorphic cameras which again are manifested from the biases which include:

(i) Contrast Sensitivity: The ability to detect changes in light intensity can be fine-tuned for different environments and applications.

(ii) Background Rate: The rate of events generated in the absence of significant changes, indicative of the sensor's noise level.

(iii) Pixel Response Time, Latency, and Jitter: These metrics describe the sensor's temporal performance, crucial for applications requiring precise timing.

\section{Proposed approach and System Design}

\begin{figure*}[ht!]
    \centering
    \includegraphics[width=0.8\textwidth]{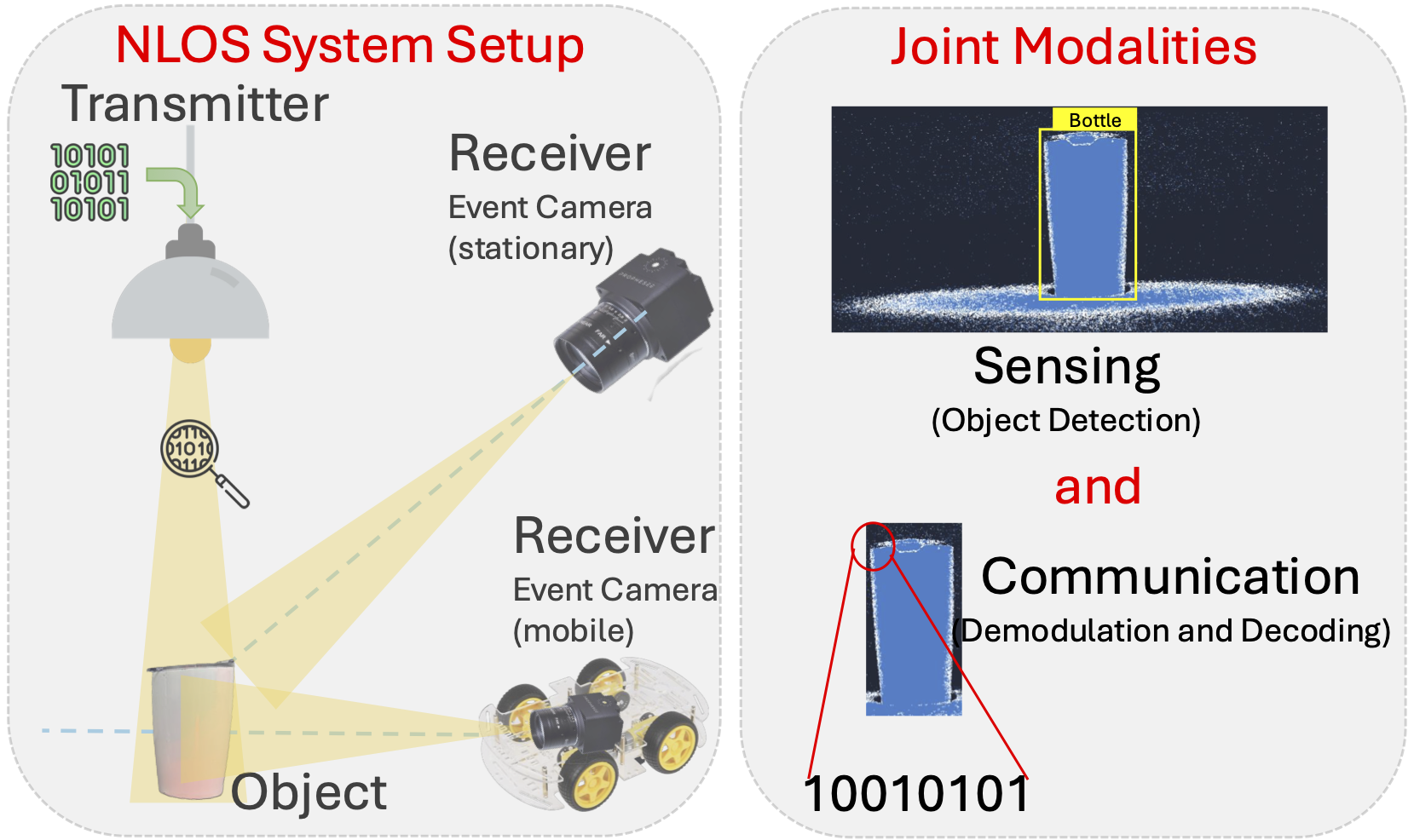}
    \caption{Conceptual illustration of the proposed joint sensing and communication using optical wireless and neuromorphic camera on a robot}
    \vspace{-4mm}
    \label{fig:concept}
\end{figure*}

We promote our concept based on the idea of a use case of a neuromorphic camera in an indoor setting where the ceiling lights (or any diffused ambient lighting) in the room are enabled for optical wireless transmission (e.g. transmit a packetized data using VLC), as shown in Figure~\ref{fig:concept}. We envision that the camera is setup for monitoring and/or detecting objects in the room, such that the camera is at an NLoS viewpoint with respect to the ceiling optical emitter. We make a case for the event camera to work in such NLoS scenarios, even when it is static (attached to the wall or immovable infrastructure) or mobile (placed on a robot performing a mission in the room. The event camera will be able to receive data streams from light emissions by demodulating the reflections from the objects, and simultaneously be able to sense (detect) objects in the vicinity. The latter is possible through spatial restructuring of the event data and executing computer vision techniques in the resultant frame.

\begin{figure*}[t!]
    \centering
\includegraphics[width=\textwidth]{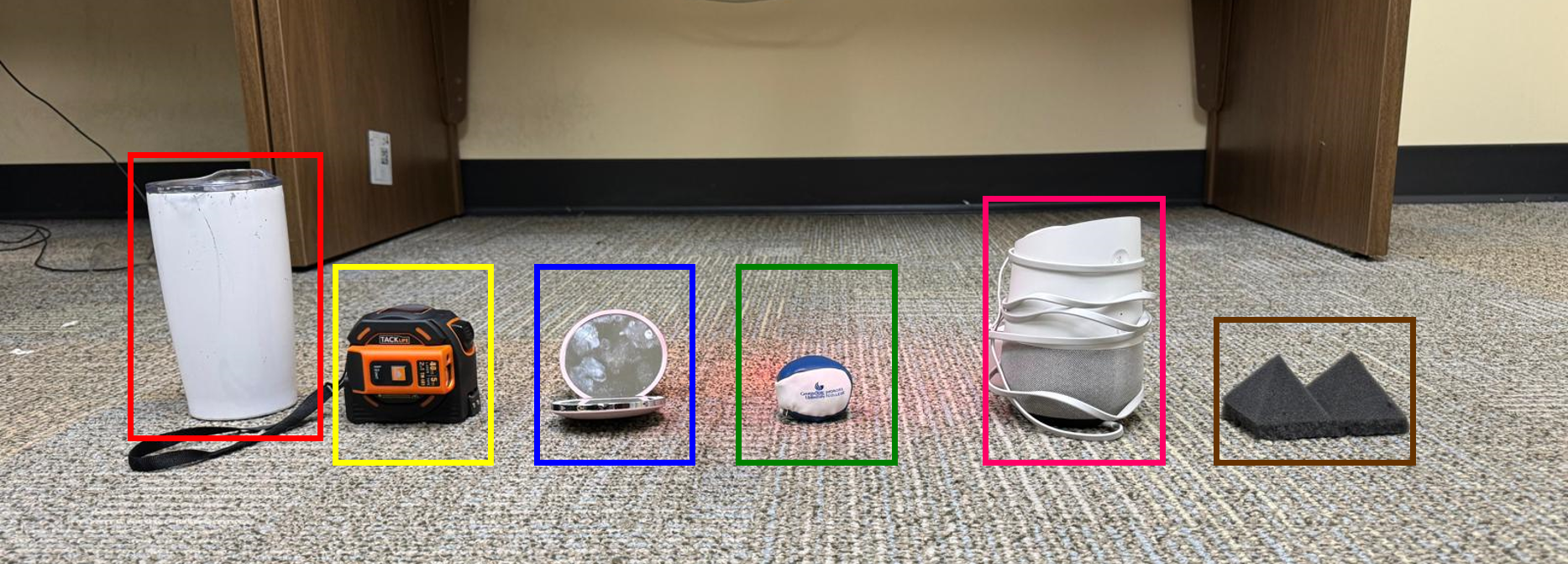}
    \caption{Various objects used in the experiment to test different surface properties and their effects in sensing and communication. From left to right - Glossy white flask (red box), Matte finish measuring tape (yellow box), Compact mirror (blue box), glossy finish leather ball (green box), Matte finish white colored google nest (magenta box), Sound reflecting foam (completely absorptive) (brown box)}
    \vspace{-2mm}
    \label{fig:objects_used}
\end{figure*}

\begin{figure*}[ht!]
    \centering
    \includegraphics[width=\textwidth]{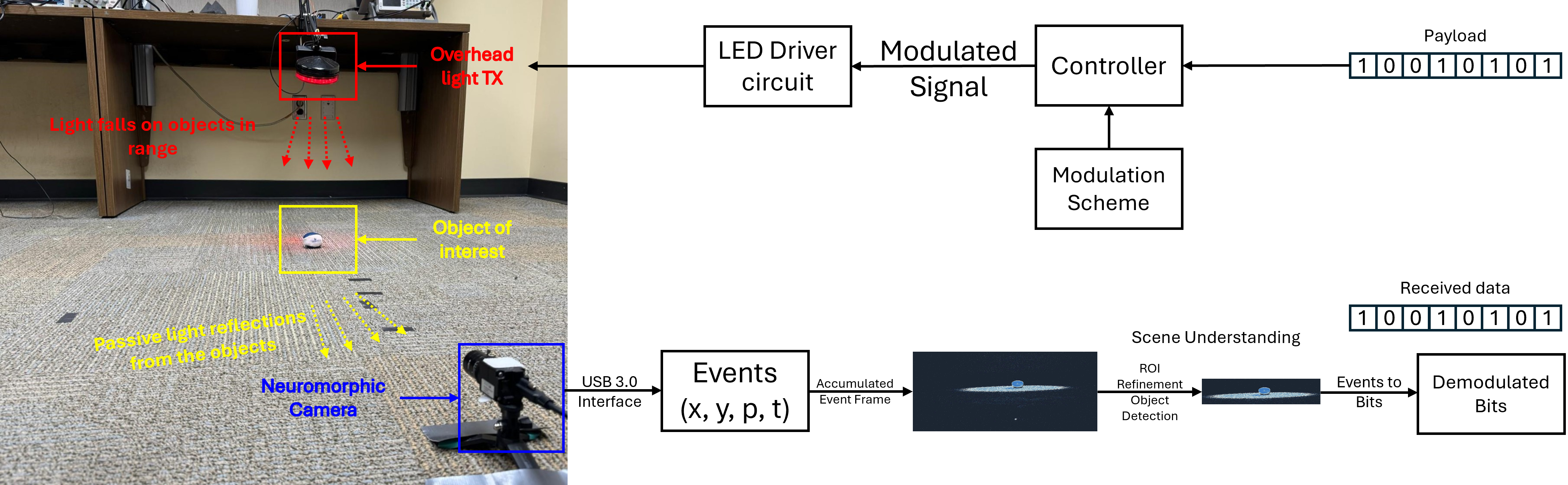}
    \caption{This figure illustrates the overall system block diagram of the receiver and its main tasks that help us achieve this modality of joint sensing and communication. Firstly, we have a neuromorphic camera that is observing the scene. Then from the USB 3.0 interface, we get the events of that scene. Then we do a periodic events accumulation framing to get the event frame representation as shown in the next step, and then we find the object of interest and region to focus on using the RoI refinement technique. Further from there we have our demodulation and decoding logic that converts all the events to bits.}
    \vspace{-4mm}
    \label{fig:block_diag}
\end{figure*}

In this work, we primarily study and demonstrate the feasibility and performance of NLoS VLC using event cameras under static conditions. This is challenging for event cameras as an event is generated when the light intensity variations occur, which only happen when there is movement of the object and/or the camera. When they are stationary, no events will get registered unless there is an active optical signal modulation involved. We wanted to characterize this modality of communication with a simple static setup where both the object of interest and the camera are not moving and see how we can achieve NLoS communication using the event camera and how these processes go hand in hand in helping each other. Figure~\ref{fig:block_diag} illustrates our envisioned setup and which will be used for our evaluations. We have an overhead light transmitter that is modulated by switching on and off at high frequencies such that it is imperceptible to the human eye. The object of interest is placed in the region of illumination of the overhead light. Then the reflections from the object of interest are captured as events by the event camera placed in the perpendicular direction of the emitter-to-object axis. 
We tested the configuration for different objects as shown in Figure~\ref{fig:objects_used}. 

Illustrated in Figure~\ref{fig:block_diag}, the neuromorphic camera captures the scene with all biases set to default values. Events are transmitted through the USB 3.0 interface of the EVK4 kit. Once received, the Periodic Frame Generation Algorithm accumulates events over a specified time and outputs them as frames, as shown in Figure~\ref{fig:block_diag}. Not all regions provide clean signals, so we refine the RoI (Region of Interest), which performs a hardware overwrite on selected pixels. This focuses the readout on a specific window, reducing background noise from other pixels.
Following that step, we tune the refractory period biases, ensuring the pixels’ dead time adheres to the Nyquist sampling theorem by setting the refractory period to $2\times F_{transmitter}$. In our case the transmitter frequency is 10Khz and we set the refractory period to 50 microseconds. Although this does not affect contrast sensitivity, it optimizes pixel availability. Finally, our algorithm performs both demodulation and decoding, recovering the data bits from the processed events. 

\section{Periodic Event Frame Generation}

The PeriodicFrameGeneration algorithm~\cite{prophesee_frames_generators} generates frames from event streams at regular intervals, ensuring that the frames are consistently spaced over time. It is particularly useful in applications requiring visualization at a fixed frequency, such as matching display or video encoder requirements. The algorithm processes all events up to the current timestamp, supporting overlapping frames, where the interval between consecutive frames can be shorter than the accumulation time. Frame generation is triggered internally and output through a callback function.

\begin{figure*}[t!]
    \centering
    \centerline{\includegraphics[width=0.7\textwidth]{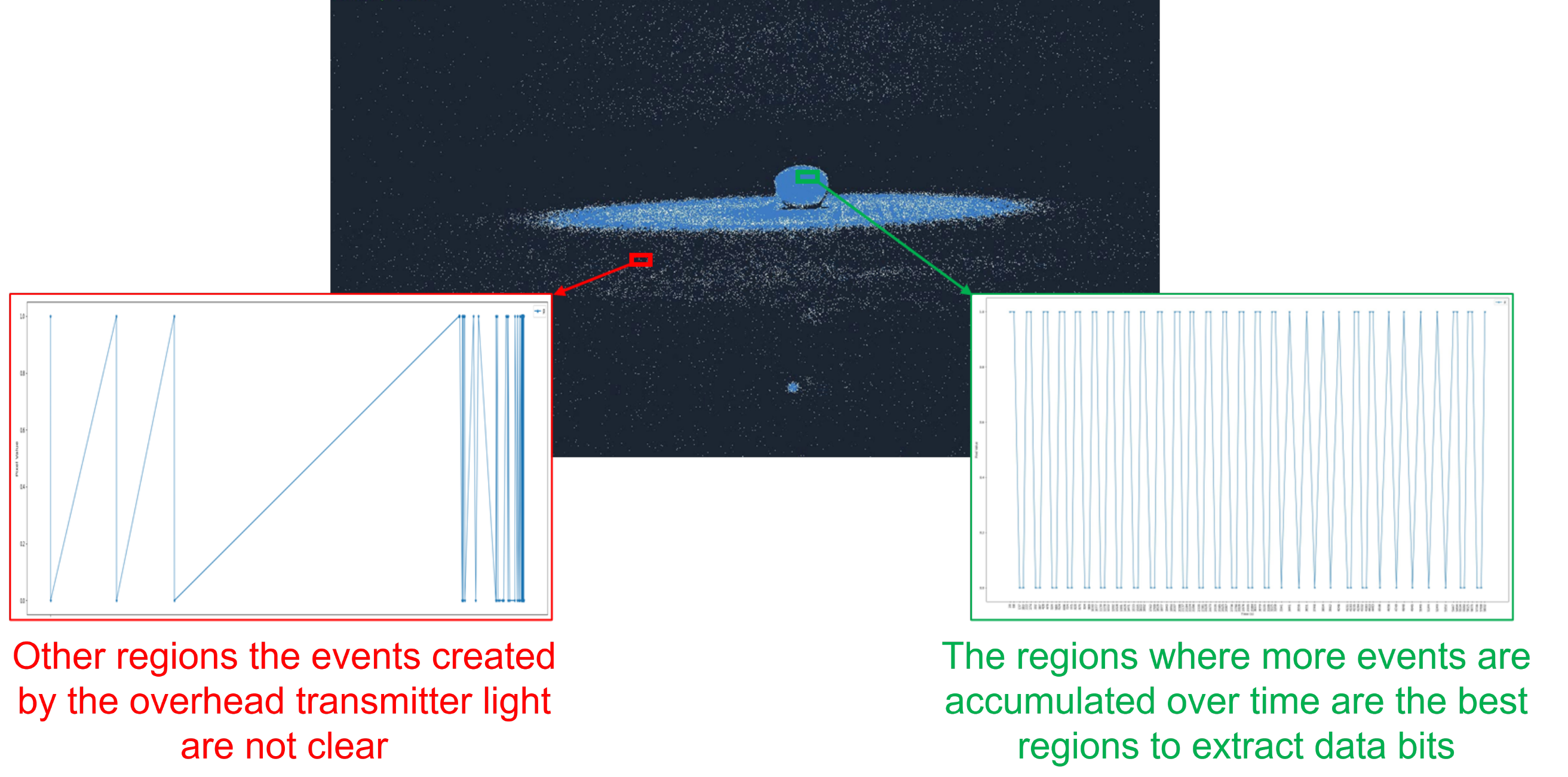}}
    \caption{The figure shows event accumulation over time in a neuromorphic camera. The green region has a higher event density, indicating optimal areas for reliable data extraction, while the red region has fewer events, reflecting weaker signals. The accompanying plots highlight the consistent event patterns in the green region compared to the red.}
    \label{fig:accum_event}
\end{figure*}

The purpose of using periodic frame generation is to help identify regions where bits can be reliably extracted. Regions with a higher accumulation of events indicate areas illuminated by overhead lights, where the scattered light reflects off the ground and surrounding objects. We observed that events accumulate primarily on objects within the illumination region of the overhead lights. Figure~\ref{fig:accum_event} demonstrates how the output of the periodic frame generation algorithm helps pinpoint regions where information can be extracted more reliably. We are transmitting simple on-off pulses from the overhead lights, and the temporal event plot from a pixel in the green region shows that the events correspond to lighting changes. In contrast, events in the red region (highlighted in Figure~\ref{fig:accum_event}) do not reflect changes in overhead lighting. 

\section{Region of Interest (RoI) Refinement}

\begin{figure}[t!]
    \centering
    \includegraphics[width=\columnwidth]{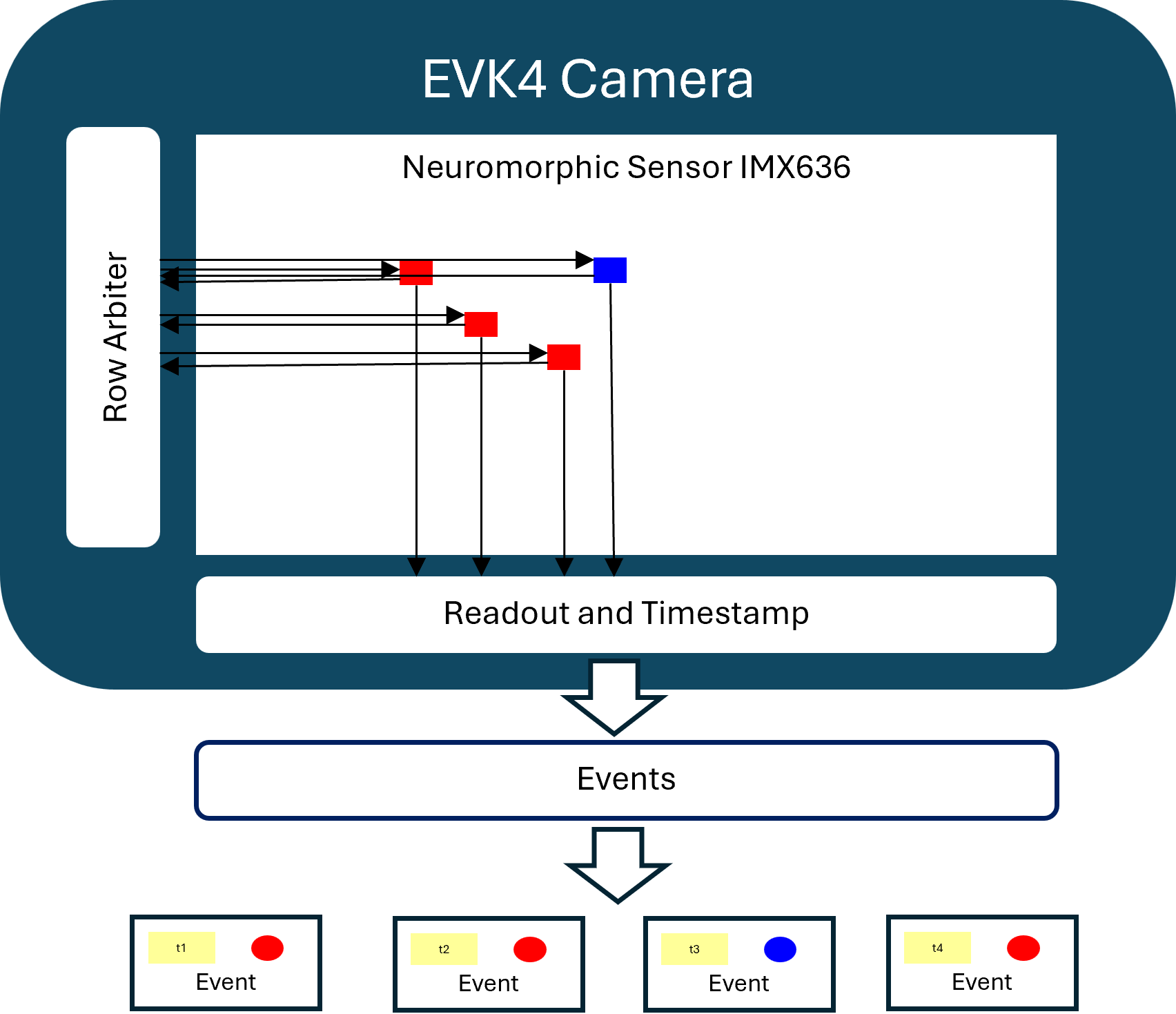}
    \caption{The figure illustrates event capture in the EVK4 camera. The row arbiter manages pixel row readouts in the IMX636 sensor, directing events to the readout and timestamp module. Each event is assigned a timestamp before output. Red and blue squares represent positive and negative polarity events, respectively.}
    \vspace{-4mm}
    \label{fig:readout_roi}
\end{figure}

Region of Interest (RoI) refinement is a very important step in our system pipeline. In the earlier sections of neuromorphic camera we talked about background rate. The background rate is the noise of the neuromorphic sensors when there are no illumination changes present in an area. If we observe Figure~\ref{fig:accum_event} we can see that there are sporadic events that accumulate in regions where the overhead light illumination is not even present in the dark scenario. This can affect our signal-to-noise ratio for our communication system. It also affects the performance of the neuromorphic camera - the way the neuromorphic camera is designed to collect events from the neuromorphic sensor is depicted in Figure~\ref{fig:readout_roi}.

\begin{figure}[t!]
    \centering
    \includegraphics[width=\columnwidth]{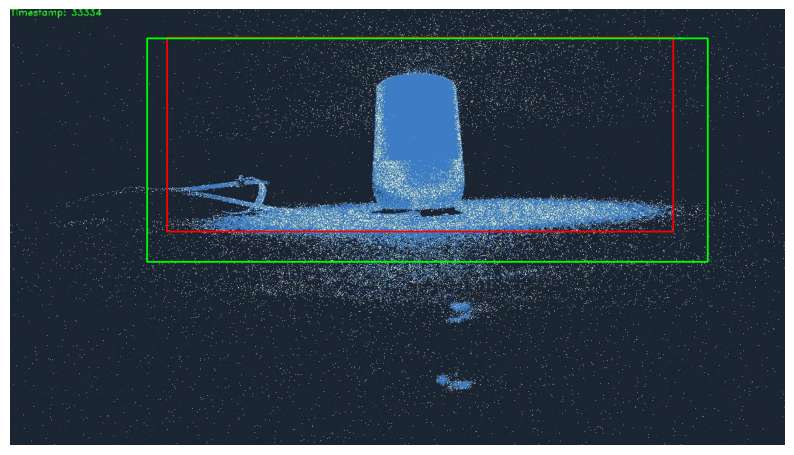}
    \caption{This figure shows the IoU evaluation where the red bounding box is the bounding box from the algorithm and green is from human annotator and then we use these boxes to calculate the IoU metric which evaluates the object presence module.}
    \vspace{-4mm}
    \label{fig:iou}
\end{figure}

In our characterization experiments, we focus on static scenarios, which contrasts with the typical applications of neuromorphic cameras in Computer Vision (CV). Neuromorphic cameras are often employed for observing and accurately tracking high-velocity objects due to their ability to capture asynchronous event streams. However, in the context of OWC, we leverage induced illumination changes to enable the use of neuromorphic cameras in a static environment. As shown in Figure~\ref{fig:accum_event}, the events generated by OWC illumination changes predominantly cluster in the region containing the object. By utilizing an accumulated event frame and applying a simple contour detection algorithm, we can localize both the object and the region of interest (RoI) for optical signal extraction.

This approach avoids the need for retraining the object detection modules provided by Prophesee, which are designed for typical neuromorphic data and do not generalize well to our OWC-specific scenarios. While more sophisticated detection methods may offer additional benefits, our use of contour detection provides a straightforward and effective solution for identifying the RoI. Specifically, we select the largest contour detected as it generally corresponds to the region containing both the object and the optical signals. The contour is then converted into a bounding box, which is used to define the RoI through the Metavision Python SDK. Finally, we record events within this RoI for decoding into bits. This naive yet effective approach allows us to reliably address the joint challenges of sensing and communication in our specific experimental setup.

\begin{table}[t!]
    \centering
    \begin{tabular}{|c|c|c|}
        \hline
        \textbf{Object} & \textbf{Average IoU} & \textbf{Max IoU} \\
        \hline
        Bottle & 0.65 & 0.70 \\
        \hline
        Mirror & 0.82 & 0.89 \\
        \hline
        Ball & 0.56 & 0.58 \\
        \hline
        Tape & 0.40 & 0.63 \\
        \hline
        Nest & 0.73 & 0.78 \\
        \hline
        Foam & 0.37 & 0.41 \\
        \hline
    \end{tabular}
    \caption{Average and Maximum IoU for Different Objects. The input event frame is first converted to a grayscale image, $I_{\text{gray}}$, to simplify the data representation. Subsequently, it is transformed into a binary image, $I_{\text{thresh}}$, using a threshold value of 50. This step enhances the contrast between the object of interest and the background, facilitating contour detection. Contours are detected in the binary image using the OpenCV \textit{findContours} function. Among the detected contours, the largest is identified by calculating the area of each contour. Finally, the largest contour is used to compute its corresponding bounding box, $B_{\text{largest}}$, which serves as the defined RoI.
}
\vspace{-10mm}
    \label{tab:iou}
\end{table}

To evaluate this approach, we use the IoU (Intersection over Union) metric, in which we label the objects on our own and calculate the IoU from the algorithm bounding box, as depicted in Figure~\ref{fig:iou}. This simple yet effective algorithm is a good baseline for object detection and RoI refinement, which is also a task that not only is crucial for reliable communication, but also jointly solves the vision task of detection of objects in a static environment. However, improvement are required in this algorithm to filter out the objects from non-objects, using heuristics or employing an ML algorithm that can filter objects. In addition, we need to employ the concept of anchor boxes to improve the object detection algorithm. Table~\ref{tab:iou} shows the results of the evaluation of the RoI refinement and object detection algorithm that we have employed; we tested it against the objects we used for the evaluation and the ground truth of the boundary boxes was obtained with human annotators. The contour detection algorithm identifies the largest contour and defines the RoI for signal collection, thereby potentially reducing background noise. In this approach, it is assumed that the largest contour corresponds to the object of interest. The algorithm processes an input event frame, $I_{\text{event}}$, and outputs the bounding box, $B_{\text{largest}}$, for the detected contour.

From Table~\ref{tab:iou}, we learn that more reflective objects have good scattering of the light, leading to illumination changes, and the foam object has the least; yet, we are able to find the presence of objects without any specialized filter design or training of a neural architecture. Another noteworthy observation is that objects that are larger in size like the bottle or nest have better IoU since we are using the logic of largest contour, which helps in better performance in those cases but a framework is in place, which can be improved iteratively. 

\section{Passive Optical Wireless Communication - From Events to Bits}
Passive optical wireless is possible due to the high dynamic range of neuromorphic cameras and also the ability to detect lights in the IR and UV spectrum, which is uncommon in other traditional camera modalities. Unlike other receivers, be it photodiode or camera, we are dealing with amplitude changes, which require manual thresholds to be set. We get direct event data, which can be translated to bits in the data. They do require tuning some biases, which is done before the communication link is established, once we have found the good parameters, we can use that during operation that reduces the need to heavy signal processing the receiver end. After we have found the RoI and recorded the events, within the RoI, we look at the pixel with the maximum number of events recorded as the data point of interest. We use the temporal data of that pixel's {\em events} to demodulate them into bits and follow up with decoding the transmitted information.

We consider two modalities of data communication:
(i) Identity (ID): Detecting an repeated ID sequence that is continuously sent by the transmitter; and
(ii) Streaming: Demodulate and decode the streaming of any arbitrary packet payload. 

\begin{figure}[t!]
    \centering
    \includegraphics[width=0.85\columnwidth]{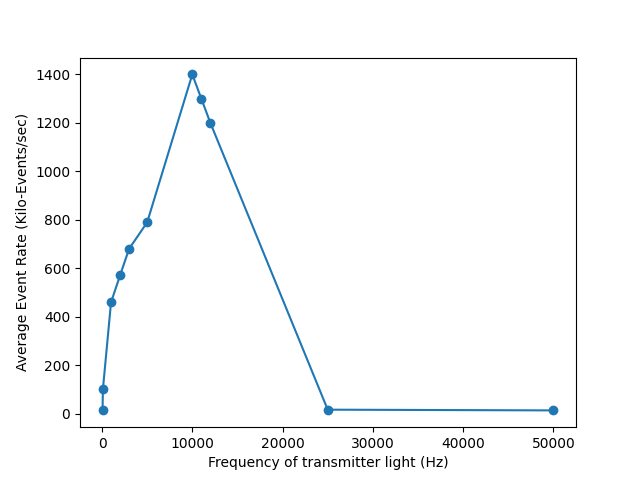}
    \caption{This figure shows the average event rate of the neuromorphic camera as we increase the frequency of switching of the transmitter. We can see that at 10KHz the average event rate increased until that point and after that it fell down which shows that it is the maximum frequency we can capture in the neuromorphic camera safely. This test was done by setting the RoI to a very small (10 x 10) pixel window.}
    \label{fig:event_freq}
\end{figure}

\vspace{2mm}\noindent{\bf Finding the highest switching frequency.} The neuromorphic camera, unlike other receivers like camera or photodiode, cannot be sampled at equal intervals, as they are asynchronous in nature. Due to this, we have to find the effective sampling rate by using a varying signal (square wave) at different frequencies.  To address this challenge, we send signals at varying frequencies and find the average events recorded per second. Using this information, we can come up with a varying curve that will tell us the effective sampling rate of our sensor.Figure~\ref{fig:event_freq} shows us that the average event rate increases until the transmitter switches to the 10KHz frequency, and after that we see that the average event rate of the camera decreases. This data was collected over a small 10 x 10 window and on the basis of this we can say that 10KHz is a maximum operating speed at which we can reliably collect data. So, our switching frequency was decided at 10KHz. 

\subsection{On Off Keying (OOK)} 

On-Off Keying (OOK) modulation is the most commonly used mechanism for pulse line coding based modulation in OWC. In OOK, a binary '1' is represented by the presence of a pulse and binary '0' is represented by the absence of the pulse.
We first test the identification (ID) detection tasks, where we have a small payload (ID bits) that is transmitted repeatedly in a loop; ID is four-bits, bit sequence of 101 and stop bit of 0. We observed that when we use simple switching we are able to effectively transmit at 10KHz and detect IDs in the dark scenario. In an ambient light scenario, this was proven to be difficult. So for the demodulation we implemented a timing-based decoder that will decode the events. Firstly, we record the data for 3 seconds and save it in a raw file (the format in which event data is stored in Metavision EVK4 devices). Then we find where the start bit has occurred which is the 101 switching pattern, then we look at the events that occurred next and appropriately demodulated to bits from events. After that, a single loop runs through the bit sequence to find the detected ID. The sliding demodulator converts the events to bits, and it takes as input the temporal event sequence of the high-frequency pixel, bit time. The bit time is the switching speed , 100KHz or 100$\mu$s equivalent time interval. 
This sliding demodulation processes a dataframe of pixel events to decode a sequence of bits based on timing information. The dataframe \textit{ df\_pixel} contains columns for the pixel values and the corresponding timestamps. The \texttt{sliding\_demodulator} algorithm is designed to process pixel events from a neuromorphic camera, stored in a DataFrame (\texttt{df\_pixel}), to demodulate and reconstruct a sequence of bits based on event timing. This method ensures accurate demodulation by detecting synchronization patterns, managing transitions between different synchronization modes, and translating consecutive patterns into corresponding bit pairs. The algorithmic pseudocode is presented in Algorithm~\ref{alg:decode_payload}.

\begin{algorithm}[t!]
\raggedright
\caption{decode\_payload}
\begin{algorithmic}
\STATE \textbf{Input:} \texttt{bits}, \texttt{start\_bits}, \texttt{stop\_bits}, \texttt{payload\_length}, \texttt{expected\_combo}
\STATE \textbf{Output:} \texttt{total\_packets}, \texttt{wrong\_packets}

\STATE \texttt{total\_packets} $\gets$ 0
\STATE \texttt{wrong\_packets} $\gets$ 0

\FOR{$i \gets 0$ \TO $\texttt{len(bits)} - \texttt{len(start\_bits)} - \texttt{payload\_length} - \texttt{len(stop\_bits)}$}
    \IF{$\texttt{start\_bits} == \texttt{bits}[i:i + \texttt{len(start\_bits)}]$ \AND $\texttt{stop\_bits} == \texttt{bits}[i + \texttt{len(start\_bits)} + \texttt{payload\_length} : i + \texttt{len(start\_bits)} + \texttt{payload\_length} + \texttt{len(stop\_bits)}]$}
        \STATE \texttt{payload} $\gets$ \texttt{bits}[i + \texttt{len(start\_bits)} : i + \texttt{len(start\_bits)} + \texttt{payload\_length}]
        \IF{$\texttt{"".join(map(str, payload))} \neq \texttt{expected\_combo}$}
            \STATE \texttt{wrong\_packets} $\gets$ \texttt{wrong\_packets} + 1
        \ENDIF
        \STATE \texttt{total\_packets} $\gets$ \texttt{total\_packets} + 1
    \ENDIF
\ENDFOR

\RETURN (\texttt{total\_packets}, \texttt{wrong\_packets})
\end{algorithmic}
\label{alg:decode_payload}
\end{algorithm}

OOK has its merits for VLC but in NLoS VLC using event cameras when the same packet (same sequence of bits) is transmitted in a continuous loop. In this way, redundancy can be leveraged to decode without synchronization. However, when streaming (continuous stream of packets with different payloads) using this method, the bit error rate (BER)  and packet error rate (PER) were too high ($>10^{-3}$ and $>60\%$, respectively), which called for a different modulation strategy.


\begin{figure}[t!]
    \centering
    \centerline{\includegraphics[width=\columnwidth]{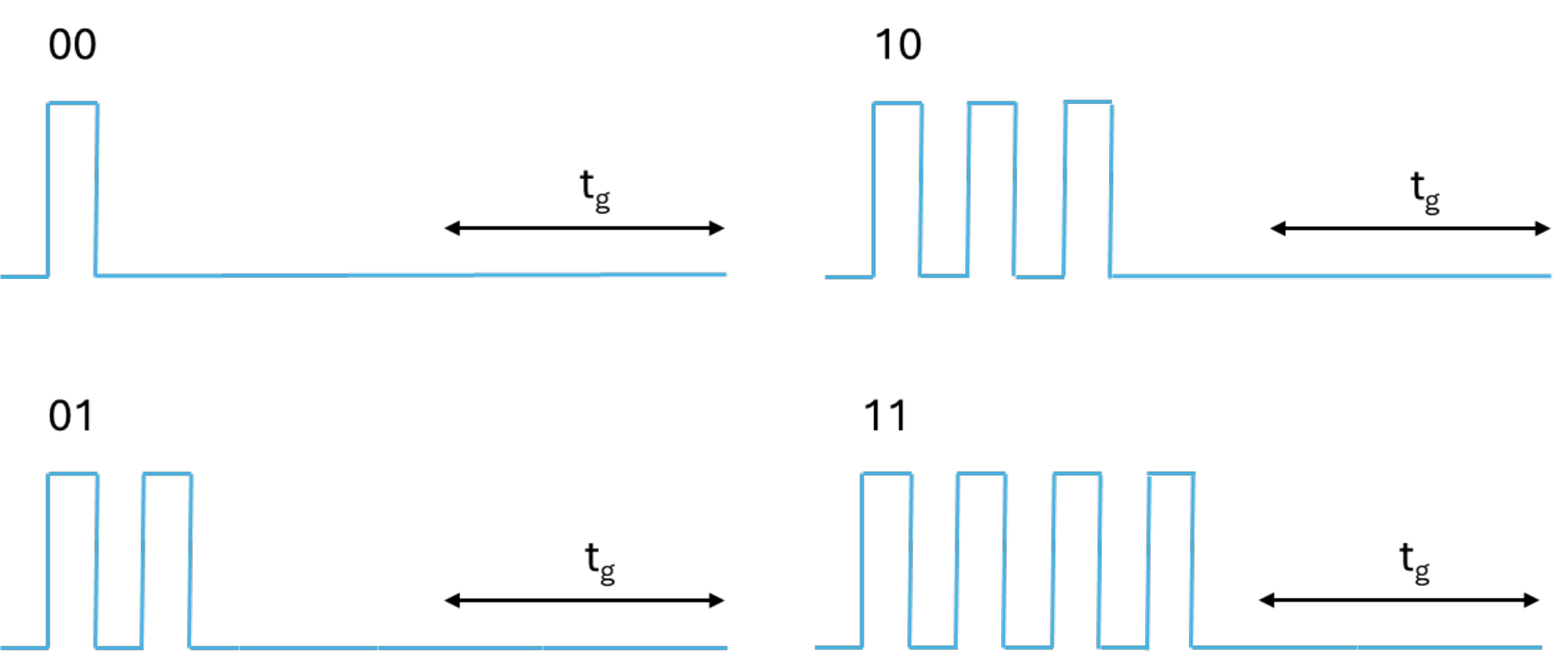}}
    \caption{This figure shows how N-pulse 4 level encoding is performed we have 4 levels from 00 to 11 encoded as pulses starting from 1 pulse to 4 pulses. Each of these symbols are followed by a guard time of 600 microseconds.}
    \vspace{-4mm}
    \label{fig:4-pulse}
\end{figure}

\begin{algorithm}[ht!]
\raggedright
\caption{Adaptive 4-level N-pulse encoding}
\begin{algorithmic}
\STATE \textbf{Function} \texttt{encode\_sequence}(\texttt{bit\_sequence})
\STATE \texttt{num\_ones} $\gets$ \texttt{count\_ones(bit\_sequence)}
\STATE \texttt{num\_zeros} $\gets$ \texttt{count\_zeros(bit\_sequence)}

\IF{\texttt{num\_ones} $>$ \texttt{num\_zeros}}
    \STATE \texttt{sync\_packet} $\gets$ \texttt{start\_bits\_11}
    \STATE \texttt{one\_one\_bit} $\gets$ \texttt{'10'} \COMMENT{Single pulse for '11'}
    \STATE \texttt{zero\_zero\_bit} $\gets$ \texttt{'10101010'} \COMMENT{Long pulse for '00'}
\ELSE
    \STATE \texttt{sync\_packet} $\gets$ \texttt{start\_bits\_00}
    \STATE \texttt{one\_one\_bit} $\gets$ \texttt{'10101010'} \COMMENT{Long pulse for '11'}
    \STATE \texttt{zero\_zero\_bit} $\gets$ \texttt{'10'} \COMMENT{Single pulse for '00'}
\ENDIF

\STATE \texttt{encoded\_sequence} $\gets$ \texttt{sync\_packet} + \texttt{guard\_bits}

\FOR{$i \gets 0$ \TO \texttt{len(bit\_sequence)} \textbf{step} 2}
    \STATE \texttt{bits} $\gets$ \texttt{bit\_sequence}[i:i+2]
    \IF{\texttt{bits} == \texttt{'11'}}
        \STATE \texttt{encoded\_sequence} $\gets$ \texttt{encoded\_sequence} + \texttt{one\_one\_bit} + \texttt{guard\_bits}
    \ELSIF{\texttt{bits} == \texttt{'10'}}
        \STATE \texttt{encoded\_sequence} $\gets$ \texttt{encoded\_sequence} + \texttt{one\_zero\_bit} + \texttt{guard\_bits}
    \ELSIF{\texttt{bits} == \texttt{'01'}}
        \STATE \texttt{encoded\_sequence} $\gets$ \texttt{encoded\_sequence} + \texttt{zero\_one\_bit} + \texttt{guard\_bits}
    \ELSE
        \STATE \texttt{encoded\_sequence} $\gets$ \texttt{encoded\_sequence} + \texttt{zero\_zero\_bit} + \texttt{guard\_bits}
    \ENDIF
\ENDFOR

\RETURN \texttt{encoded\_sequence}
\end{algorithmic}
\label{alg:n-pulse_adap}
\end{algorithm}

\subsection{N-pulse modulation -- traditional and adaptive}

The OOK version works well with the ID recognition task, where we know what bit patterns we are looking for, and we can use a correlation decoder. This fails when the pattern is unknown or there is no pattern. To this end, we build an N-pulse modulation based approach for our system, inspired by the work from Aranda et al.,~\cite{npulsemod}.

We employed a 2-symbol N-pulse scheme, where a \textbf{0} bit is encoded as a single pulse and a \textbf{1} bit as two pulses. To ensure reliable communication—especially in our overhead VLC transmitter scenario with irregular reflections from various surfaces—we introduced guard times between pulses. The chosen guard time was 600~$\mu$s.
For synchronization and start bits, we transmit a sequence of 5 pulses consecutively (i.e., without interleaved guard times), followed by a single 600~$\mu$s guard time.
Assuming that the payload exhibits an equal number of \textbf{0}’s and \textbf{1}’s, the average number of pulses per payload bit is 1.5. Given that each pulse slot occupies
\[
100~\mu\text{s} \; (\text{pulse duration}) + 600~\mu\text{s} \; (\text{guard time}) = 700~\mu\text{s},
\]
the average time to send one payload bit is
\[
1.5 \times 700~\mu\text{s} = 1050~\mu\text{s}.
\]
This corresponds to a payload bit rate of approximately
\[
\frac{1}{1050 \times 10^{-6}} \approx 952~\text{bps},
\]
ignoring the synchronization overhead (assuming synchronization is performed very sparingly).

Our packet consists of 64 bits and contains all 4-bit combinations from \texttt{0000} to \texttt{1111}. In a dark room scenario, for 293 transmitted packets, the packet error rate (PER) was 23\% and the bit error rate (BER) was $2.7\times10^{-3}$. Under ambient lighting conditions (with the room ceiling light on), the PER increased to 47\% while the BER was $4.5\times10^{-3}$.

Although the PER is relatively high, the BER remains acceptable. This indicates that, with further tuning, this mode of communication holds significant potential for practical applications. 

This gave us the intuition that the modulation of N-pulse might work for our system, as it uses a burst of pulses as start bits or synchronization bits and then uses a 4 symbol encoding. The `00' data are encoded as single pulses, `01' is encoded as two pulses, `10' is encoded as three pulses, and `11' is encoded as four pulses and then use guard time where they are not operating in the channel. The utilization of this method, we are losing bandwidth, but then this provides a robust approach to communicate in our complex channel. Our tests showed that this communication approach is more reliable. We used 10 KHz pulses and our effective data rate now assuming uniformly distributed data ignoring synchronization overhead, the average number of pulses per 2-bit symbol is 
\[
\frac{1+2+3+4}{4} = 2.5 \text{ pulses per symbol}.
\]
Since we use 10~kHz pulses, each pulse lasts 
\[
\frac{1}{10\,000} = 100\,\mu\text{s},
\]
so the burst duration per symbol is 
\[
2.5 \times 100\,\mu\text{s} = 250\,\mu\text{s}.
\]
After the burst, a guard time of 600~$\mu$s is applied, yielding a total symbol duration of 
\[
250\,\mu\text{s} + 600\,\mu\text{s} = 850\,\mu\text{s}.
\]
Since each symbol carries 2 bits, the effective data rate is 
\[
\frac{2\text{ bits}}{850 \times 10^{-6}\text{ s}} \approx 2353~\text{bps}.
\]

Our tests showed that this communication approach is more reliable. Figure~\ref{fig:4-pulse} illustrates how the n-pulse encoding is performed.

\begin{algorithm}[ht!]
\raggedright
\caption{Demodulate Bits from Indices}
\begin{algorithmic}
\STATE \textbf{Input:} \texttt{indices}, \texttt{bits}, \texttt{pattern\_length}, \texttt{sync\_threshold\_00}, \texttt{sync\_threshold\_11}
\STATE \textbf{Output:} \texttt{packets}

\STATE \texttt{decoded\_bits} $\gets$ []
\STATE \texttt{packets} $\gets$ []
\STATE \texttt{i} $\gets$ 0
\STATE \texttt{in\_sync\_00}, \texttt{in\_sync\_11} $\gets$ \textbf{false}, \textbf{false}

\WHILE{\texttt{i} $<$ \texttt{len(indices)}}
    \STATE \texttt{count} $\gets$ 1
    \WHILE{\texttt{i + count} $<$ \texttt{len(indices)} \textbf{and} \texttt{indices[i + count] == indices[i] + count * pattern\_length}}
        \STATE \texttt{count} $\gets$ \texttt{count} + 1
    \ENDWHILE
    
    \IF{\texttt{count} $\geq$ \texttt{sync\_threshold\_11} \textbf{or} \texttt{count} $\geq$ \texttt{sync\_threshold\_00}}
        \IF{\texttt{decoded\_bits} $\neq$ []}
            \STATE \texttt{packets.append(decoded\_bits)}
        \ENDIF
        \STATE \texttt{decoded\_bits} $\gets$ []
        \STATE \texttt{in\_sync\_11}, \texttt{in\_sync\_00} $\gets$ (\texttt{count} $\geq$ \texttt{sync\_threshold\_11}), (\texttt{count} $\geq$ \texttt{sync\_threshold\_00})
        \STATE \texttt{i} $\gets$ \texttt{i + count}
        \STATE \textbf{continue}
    \ENDIF
    
    \IF{\texttt{in\_sync\_11}}
        \STATE \texttt{decoded\_bits.extend(\{1:[1,1], 2:[0,1], 3:[1,0], 4:[0,0]\}.get(count, []))}
    \ELSIF{\texttt{in\_sync\_00}}
        \STATE \texttt{decoded\_bits.extend(\{1:[0,0], 2:[0,1], 3:[1,0], 4:[1,1]\}.get(count, []))}
    \ENDIF

    \STATE \texttt{i} $\gets$ \texttt{i + count}
\ENDWHILE

\IF{\texttt{decoded\_bits} $\neq$ []}
    \STATE \texttt{packets.append(decoded\_bits)}
\ENDIF

\RETURN \texttt{packets}
\end{algorithmic}
\label{alg:bits_indices}
\end{algorithm}

To improve this further, we propose a new mechanism, called {\bf N-pulse adaptive}. In our new mechanism we observe the individual packets, if the number of zeros in the packets are more we use the traditional n-pulse as described above otherwise if the number of ones in the packet is higher than '11' bits become one pulse and then '00' bits become four pulses symbol. This is communicated using the synchronization bits, where if we send 8 pulses we use the default scheme, and if we send 11 pulses in synchronization we use the second scheme. The n pulse adaptive algorithm is represented in Algorithm~\ref{alg:n-pulse_adap}.
The decoder works for this modulation uses the sliding demodulator from earlier to get the bits and then we use a '1-0' detector which returns the indices of '1-0' occurring in the data, then we use the indices to decode the bits based on how many '1-0' occur close to each other and we use that information to decode our bits which is shown in Algorithm~\ref{alg:bits_indices}.

\section{Evaluation}

In this section, we discuss the evaluation of our system. We first examine OOK modulation in terms of detecting ID sequences, and its shortcomings in streaming arbitrary packets will be highlighted. We then delve into N-pulse modulation and its adaptive variant for arbitrary packet transmission. All evaluations were conducted with the event camera pointed at the object parallel to the ground at a distance of 1 meter and the emitter shining on the object in the top-down view from a height of 1 meter. 
Although this static setup allows precise analysis of the modulation schemes, dynamic scenarios may introduce additional challenges such as temporal de-synchronization and increased error rates. Future work will address these challenges by investigating adaptive error correction and synchronization mechanisms for dynamic environments.

\subsection{OOK Modulation and ID Detection}

\begin{figure}[t!]
    \centering
    \includegraphics[width=\columnwidth]{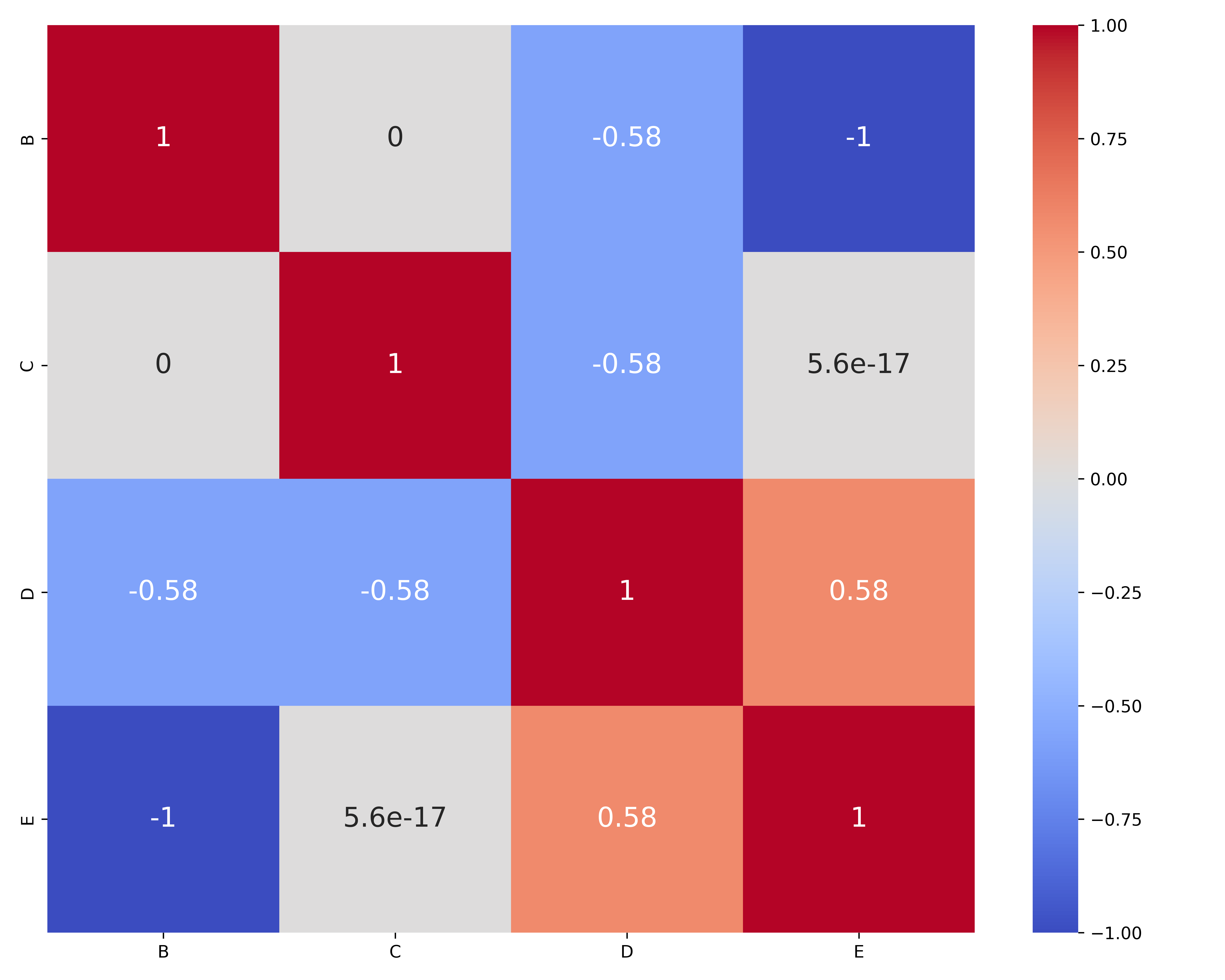}
    \caption{Pearson correlation coefficients of 4 bit sequences chosen for ID detection. This shows that they are not correlated and are a good choice to use together.}
    \vspace{-4mm}
    \label{fig:corr}
\end{figure}

We tested ID detection with OOK modulation and successfully differentiated between 6 distinct 4-bit IDs that had 101 as start bits and 0 as stop bits. Of all possible 4-bit sequences, 6 demonstrated low correlation, as depicted in Figure~\ref{fig:corr}. Note that sequences 0000 (A) and 1111 (F) were not used due to their zero standard deviation, making the Pearson correlation undefined. Thus, these sequences were selected as the IDs to detect. The switching frequency was set at 10 kHz for ID detection.

\begin{table}[t!]
\centering
\scalebox{0.8}{
\begin{tabular}{|c|c|c|c|}
    \hline
    \textbf{Label} & \textbf{Bit Combination} & \textbf{Packet Error Rate} & \textbf{Total Packets} \\
    \hline
    A & 0000 & 5\% & 129 \\
    \hline
    B & 0011 & 17\% & 3395 \\
    \hline
    C & 0101 & 2\% & 22014 \\
    \hline
    D & 1000 & 1\% & 467 \\
    \hline
    E & 1100 & 1\% & 3907 \\
    \hline
    F & 1111 & 1\% & 5675 \\
    \hline
\end{tabular}}
\caption{Packet error rates and total packets for specified bit combinations}
\label{tab:ookatof}
\end{table}

\begin{table*}[h!]
\centering
\scalebox{0.9}{
\begin{tabular}{|c|c|c|c|c|c|}
\cline{2-6}
\multicolumn{1}{r|}{}  & \textbf{Packet Error Rate} & \textbf{Total Packets} & \textbf{Avg. Hamming Distance} & \textbf{Max. Hamming Distance} & \textbf{Bit Error Rate} \\ \hline
\textbf{Dark} & 2\% & 233 & 0.03 & 2 & $4.02 \times 10^{-4}$ \\ \hline
\textbf{Ambient} & 17\% & 217 & 0.97 & 20 & $1.46 \times 10^{-2}$ \\ \hline
\end{tabular}}
\caption{Traditional N-pulse 4-level results in both ambient and dark scenarios}
\label{table:4-pulse_results}
\end{table*}

\begin{table*}[!htbp]
\centering
\scalebox{0.85}{
\begin{tabular}{c|c|c|c|c|c|}
\cline{2-6}
\multicolumn{1}{r|}{} & \textbf{Packet Error Rate} & \textbf{Total Packets} & \textbf{Avg. Hamming Distance} & \textbf{Max. Hamming Distance} & \textbf{Bit Error Rate} \\
\hline
\multicolumn{1}{|c|}{\textbf{Dark (equal)}} & 2\% & 233 & 0.03 & 2 & $4.02 \times 10^{-4}$ \\
\hline
\multicolumn{1}{|c|}{\textbf{Ambient (equal)}} & 11\% & 644 & 0.97 & 35 & $1.47 \times 10^{-2}$ \\
\hline
\multicolumn{1}{|c|}{\textbf{Dark (1>0)}} & 1\% & 211 & 0.02 & 2 & $3.75 \times 10^{-4}$ \\
\hline
\multicolumn{1}{|c|}{\textbf{Ambient (1>0)}} & 7\% & 322 & 0.85 & 32 & $1.46 \times 10^{-2}$ \\
\hline
\end{tabular}}
\caption{N-Pulse adaptive results in ambient and dark conditions, examining sequences with equal numbers of zeros and ones, and sequences with more ones than zeros (1>0).}
\label{table:resultsnadap}
\end{table*}

\begin{table*}[ht!]
\centering
\scalebox{0.85}{
\begin{tabular}{c|c|c|c|c|c|}
\cline{2-6}
\multicolumn{1}{r|}{} & \textbf{Packet Error Rate} & \textbf{Total Packets} & \textbf{Avg. Hamming Distance} & \textbf{Max. Hamming Distance} & \textbf{Bit Error Rate} \\
\hline
\multicolumn{1}{|c|}{\textbf{Mirror (Dark equal)}} & 1\% & 130 & 0 & 1 & $1 \times 10^{-11}$  \\
\hline
\multicolumn{1}{|c|}{\textbf{Mirror (Ambient equal)}} & 20\% & 213 & 2.43 & 31 & $2.43 \times 10^{-2}$ \\
\hline
\multicolumn{1}{|c|}{\textbf{Mirror (1>0 Dark)}} & 0.3\% (1 packet) & 343 & 0 & 1 & $1 \times 10^{-11}$ \\
\hline
\multicolumn{1}{|c|}{\textbf{Mirror (1>0 Ambient)}} & 18.4\% & 223 & 2.85 & 28 & $1.98 \times 10^{-2}$ \\
\hline
\multicolumn{1}{|c|}{\textbf{Ball (Dark Equal)}} & 2\% & 233 & 0.03 & 2 & $4.02\times 10^{-4}$ \\
\hline
\multicolumn{1}{|c|}{\textbf{Ball (Ambient equal)}} & 11\% & 644 & 0.97 & 35 & $1.47 \times 10^{-2}$ \\
\hline
\multicolumn{1}{|c|}{\textbf{Ball (1>0 dark)}} & 1\% & 211 & 0.02 & 2 & $3.75 \times 10^{-4}$ \\
\hline
\multicolumn{1}{|c|}{\textbf{Ball (1>0 ambient)}} & 7\% & 322 & 0.85 & 32 & $1.46 \times 10^{-2}$ \\
\hline
\multicolumn{1}{|c|}{\textbf{Google Nest (Dark equal)}} & 33\% & 145 & 0.53 & 3 & $8.32 \times 10^{-3}$ \\
\hline
\multicolumn{1}{|c|}{\textbf{Flask (Dark equal)}} & 35\% & 153 & 0.59 & 4 & $9.83 \times 10^{-3}$ \\
\hline
\multicolumn{1}{|c|}{\textbf{Tape (Dark equal)}} & 100\% & 263 & 31.23 & 35 & $4.78 \times 10^{-1}$ \\
\hline
\multicolumn{1}{|c|}{\textbf{Foam (Dark equal)}} & \multicolumn{5}{c|}{ No packets detected} \\
\hline
\end{tabular}}
\caption{N-Pulse adaptive 4-level modulation results for various objects}
\label{table:results_objects}

\end{table*}

We consistently recorded the PER for the sequences A-F, which was in the order of 30\%, and higher for the streaming arbitrary packets case.
However, the data in Table~\ref{tab:ookatof} shows that OOK modulation is generally suitable for detecting ID-based events.

\subsection{Traditional N-Pulse Modulation}
All experiments were performed on the ball baseline object. The effective bit rate for the 4-level N-pulse modulation is 1454.54 bps.
From Table~\ref{table:4-pulse_results}, we see that the system has reasonable BERs and PERs for reliable communication. However the data rate needs improvement.

\subsection{N-Pulse Adaptive}

The adaptive N-pulse modulation has three cases for data rate:
\begin{itemize}
    \item \textbf{Best case:} When the packet consists entirely of 0s or 1s, the data rate is 1828.57 bps.
    \item \textbf{Average case:} When the packet has a slightly greater number of 1s than 0s or vice versa, the data rate is 1702.13 bps.
    \item \textbf{Worst case:} When the packet has an equal number of 0s and 1s, the data rate is 1454.54 bps.
\end{itemize}

These estimates from the packets we created for transmission including the synchronization.
The results in Table~\ref{table:resultsnadap} provide a comprehensive overview of the performance of the N-pulse adaptive modulation scheme under different lighting conditions and bit sequence compositions. In dark conditions, the N-Pulse adaptive modulation scheme shows remarkable performance. When the bit sequences have an equal number of zeros and ones, the Packet Error Rate is 0.02, which is an average hamming distance of 0.03 and a maximum hamming distance of 2. The Bit Error Rate (BER) is 0.000402. This indicates that the system is reliable in transmitting data with minimal errors in a controlled, low-light environment. 

When the bit sequences have more ones than zeros, the performance improved but is not that significant. This shows that the adaptive modulation scheme effectively leverages the bit composition to enhance reliability. 

Under ambient lighting conditions, the performance of the N-pulse adaptive modulation scheme degrades compared to the dark scenarios, yet it still remains relatively robust. For bit sequences with an equal number of zeros and ones, the PER increases to 0.11 and the BER to 0.015. The average Hamming distance increases by 0.97 and the maximum Hamming distance reaches 35. These results suggest that ambient light introduces more noise and potential interference, leading to higher error rates. In sequences where there are more ones than zeros, the system slightly performs better than equal bit sequences under ambient light. This indicates that the adaptive modulation scheme can partially mitigate the impact of ambient light by adjusting the encoding based on bit composition. 

Despite the challenges posed by ambient lighting, the system maintains relatively low error rates. The ability to keep the PER and BER within acceptable limits under different conditions highlights the robustness of the proposed modulation scheme. Furthermore, Hamming distance metrics provide additional insight into the nature of errors. The lower average Hamming distance under dark conditions indicates fewer bit flips, while higher distances under ambient conditions highlight the increased error likelihood. The maximum Hamming distance observed under ambient conditions points to occasional severe errors, likely due to sudden changes in lighting or reflections.

\subsection{N-Pulse Adaptive for Different Objects}


We evaluated the N-pulse adaptive modulation for various daily objects, as shown in Figure~\ref{fig:objects_used}.
From Table~\ref{table:results_objects}, we observe that the system performs well with mirror and ball, which have glossy finishes. In contrast, larger objects such as the bottle and Google Nest, which are closer to the light source, suffer from higher error rates due to multipath reflections. Matte finish objects also exhibit high error rates, and the foam, being completely absorptive, did not detect any packets. The extended results for these experiments are presented in Table~\ref{table:results_objects}. We derive the following insights from these results:

\vspace{1mm}\noindent\textbf{Impact of Ambient Light}: Objects tested under ambient light conditions generally show a significant increase in PER and BER, with longer average and maximum Hamming distances. This indicates that ambient light introduces noise and errors into transmission, degrading performance. 

\vspace{1mm}\noindent\textbf{Material Properties:} The performance varies significantly with different objects. These are also interpreted above in detail where the reflectivity and color of the objects play an important role. Additionally, we also observe that perfect reflective surfaces like mirrors have higher degraded performance under ambient conditions, whereas objects like foam fail to detect any packets. The variation suggests that material properties play a crucial role in our communication system;
(i) lighter colored objects, which are more reflective, perform better than darker colored objects;
(ii) objects with glossy finishes outperform those with matte finishes;
(iii) the size of the objects and their proximity to the light source matter significantly (short objects leverage the light diffusion  minimizing multipath effects).



\section{Conclusion}
In this paper, we explored the potential of neuromorphic cameras for non-line-of-sight optical wireless communication. Our evaluation demonstrates the feasibility of using Off-Off Keying under high redundancy, but it falls short of streaming arbitrary packets due to high error rates. Traditional N-pulse modulation shows promise with practically bearable packet error rates (PER) and bit error rates (BER) under both dark and ambient conditions. The adaptive N-pulse modulation further improves reliability by dynamically adjusting the encoding scheme based on the bit composition of the packets, achieving higher data rates in various scenarios.
We also evaluated the system's performance with different daily objects, highlighting that lighter-colored and glossy-finish objects, which are more reflective, exhibit better performance. Larger objects and those with matte finishes tend to suffer from higher error rates, particularly in the presence of multipath reflections.

Our findings indicate feasibility and practicality of using neuromorphic cameras for passive optical wireless communication, leveraging their high dynamic range and sensitivity to changes in illumination. Future work will focus on enhancing data rates and refining object detection techniques to further improve the system's sensing and communication capabilities. As an extension of this work in future, we plan to address such mobility challenges by investigating adaptive error correction and synchronization mechanisms for dynamic environments.

\section{Acknowledgments}
We thank the anonymous reviewers and the shepherd for their feedback and constructive suggestions to improve this paper. This work was supported by the National Science Foundation under
grant number CNS-2146267.

\balance
\bibliographystyle{unsrt}
\bibliography{references}

\begin{thebibliography}{10}

\bibitem{2024spikingneuralnetworksfastmoving}
Andreas Ziegler, Karl Vetter, Thomas Gossard, Jonas Tebbe, and Andreas Zell.
\newblock Spiking neural networks for fast-moving object detection on neuromorphic hardware devices using an event-based camera, 2024.

\bibitem{neuro}
Terrence Stewart, Marc-Antoine Drouin, Michel Picard, Frank Billy~Djupkep, Anthony Orth, and Guillaume Gagn\'{e}.
\newblock Using neuromorphic cameras to track quadcopters.
\newblock In {\em Proceedings of the 2023 International Conference on Neuromorphic Systems}, ICONS '23, New York, NY, USA, 2023. Association for Computing Machinery.

\bibitem{meetoptics_photodiodes}
{MeetOptics}.
\newblock Photodiodes - what is a photodiode?
\newblock \url{https://www.meetoptics.com/academy/photodiodes#what-is-a-photodiode}, Accessed: 2024-06-29.

\bibitem{aranda2020npulse}
Juan Aranda, Hakyong Kim, and Alireza Aghasi.
\newblock n-pulse modulation for event-based optical camera communication.
\newblock {\em IEEE Transactions on Signal Processing}, 68:1234--1245, 2020.

\bibitem{fang2022joint}
Junyi Fang, Yutao Li, Shiyang Han, and Ling Xiao.
\newblock Joint communication and sensing in 6g: A survey.
\newblock {\em IEEE Journal on Selected Areas in Communications}, 40(3):1620--1633, 2022.

\bibitem{hrm}
Yu~Gu, Xiang Zhang, Zhi Liu, and Fuji Ren.
\newblock Wifi-based real-time breathing and heart rate monitoring during sleep, 2019.

\bibitem{deng2023occreview}
Wei Deng, Xian Liu, and Weiyu Liu.
\newblock A comprehensive review of optical camera communication systems.
\newblock {\em IEEE Access}, 11:105--124, 2023.

\bibitem{vlc_backscat}
Kenuo Xu, Kexing Zhou, Chengxuan Zhu, Shanghang Zhang, Boxin Shi, Xiaoqiang Li, Tiejun Huang, and Chenren Xu.
\newblock When visible light (backscatter) communication meets neuromorphic cameras in v2x.
\newblock In {\em Proceedings of the 24th International Workshop on Mobile Computing Systems and Applications}, HotMobile '23, page 42–48, New York, NY, USA, 2023. Association for Computing Machinery.

\bibitem{remark}
Tzu-Hsu Yu and Hsin-Mu Tsai.
\newblock {\em ReMark: Privacy-preserving Fiducial Marker System via Single-pixel Imaging}.
\newblock Association for Computing Machinery, New York, NY, USA, 2023.

\bibitem{wang2024highspeedpassivevisiblelight}
Yanxiang Wang, Yiran Shen, Kenuo Xu, Guangrong Zhao, Mahbub Hassan, Chenren Xu, and Wen Hu.
\newblock Towards high-speed passive visible light communication with event cameras and digital micro-mirrors, 2024.

\bibitem{gehrig2022eventbased}
Daniel Gehrig, Henri Rebecq, Guillermo Gallego, and Davide Scaramuzza.
\newblock Event-based vision: A survey.
\newblock {\em IEEE Transactions on Pattern Analysis and Machine Intelligence}, 44(1):154--171, 2022.

\bibitem{ziegler2024spiking}
Andreas Ziegler, Karl Vetter, Thomas Gossard, Jonas Tebbe, and Andreas Zell.
\newblock Spiking neural networks for fast-moving object detection on neuromorphic hardware devices using an event-based camera, 2024.

\bibitem{prophesee_evk4}
Prophesee.
\newblock {\em EVK4 Event-Based Vision Kit}, 2024.
\newblock Accessed: 2024-07-02.

\bibitem{prophesee_frames_generators}
Prophesee.
\newblock {\em Prophesee Documentation: Frames Generators Guide}, 2024.
\newblock Accessed: 2024-07-02.

\bibitem{npulsemod}
Jaime Aranda, Victor Guerra, Jose Rabadan, and Rafael Perez-Jimenez.
\newblock Enhancing computational efficiency in event-based optical camera communication using n-pulse modulation.
\newblock {\em Electronics}, 13(6), 2024.

\end{thebibliography}

\end{document}